\documentclass[10pt,twocolumn,letterpaper]{article}

\usepackage{iccv}              
\usepackage{multirow}

\definecolor{iccvblue}{rgb}{0.21,0.49,0.74}
\usepackage[pagebackref,breaklinks,colorlinks,allcolors=iccvblue]{hyperref}
\usepackage[linesnumbered,ruled,vlined]{algorithm2e}

\def\paperID{1542} 
\def\confName{ICCV}
\def\confYear{2025}

\title{LaneDiffusion: Improving Centerline Graph Learning via Prior Injected BEV Feature Generation}

\author{
    Zijie Wang$^{1,2}$\thanks{Equal contribution. Work done during an internship at Baidu.},
    Weiming Zhang$^{3}$\footnotemark[1],
    Wei Zhang$^{3}$\footnotemark[1],
    Xiao Tan$^{3}$,
    Hongxing Liu$^{3}$,
    Yaowei Wang$^{4,5}$,
    Guanbin Li$^{1,2,5,6}$\thanks{Corresponding author.}\\[1ex]
    $^1$Sun Yat-sen University, $^2$Shenzhen Loop Area Institute, $^3$Baidu Inc.\\
    $^4$Harbin Institute of Technology, Shenzhen, $^5$Pengcheng Laboratory\\
    $^6$Guangdong Key Laboratory of Big Data Analysis and Processing \\
    \small
    \texttt{wangzj75@mail2.sysu.edu.cn, liguanbin@mail.sysu.edu.cn}\\
}

\begin{document}
\maketitle
\begin{abstract}
    Centerline graphs, crucial for path planning in autonomous driving, are traditionally learned using deterministic methods. However, these methods often lack spatial reasoning and struggle with occluded or invisible centerlines. Generative approaches, despite their potential, remain underexplored in this domain. We introduce LaneDiffusion, a novel generative paradigm for centerline graph learning. LaneDiffusion innovatively employs diffusion models to generate lane centerline priors at the Bird's Eye View (BEV) feature level, instead of directly predicting vectorized centerlines. Our method integrates a Lane Prior Injection Module (LPIM) and a Lane Prior Diffusion Module (LPDM) to effectively construct diffusion targets and manage the diffusion process. Furthermore, vectorized centerlines and topologies are then decoded from these prior-injected BEV features. Extensive evaluations on the nuScenes and Argoverse2 datasets demonstrate that LaneDiffusion significantly outperforms existing methods, achieving improvements of $4.2\%$, $4.6\%$, $4.7\%$, $6.4\%$ and $1.8\%$ on fine-grained point-level metrics (GEO F1, TOPO F1, JTOPO F1, APLS and SDA) and $2.3\%$, $6.4\%$, $6.8\%$ and $2.1\%$ on segment-level metrics (IoU, mAP$_{cf}$, DET$_{l}$ and TOP$_{ll}$). These results establish state-of-the-art performance in centerline graph learning, offering new insights into generative models for this task.
\end{abstract}

\begin{figure}[!ht]
    \centering
    \includegraphics[width=\linewidth]{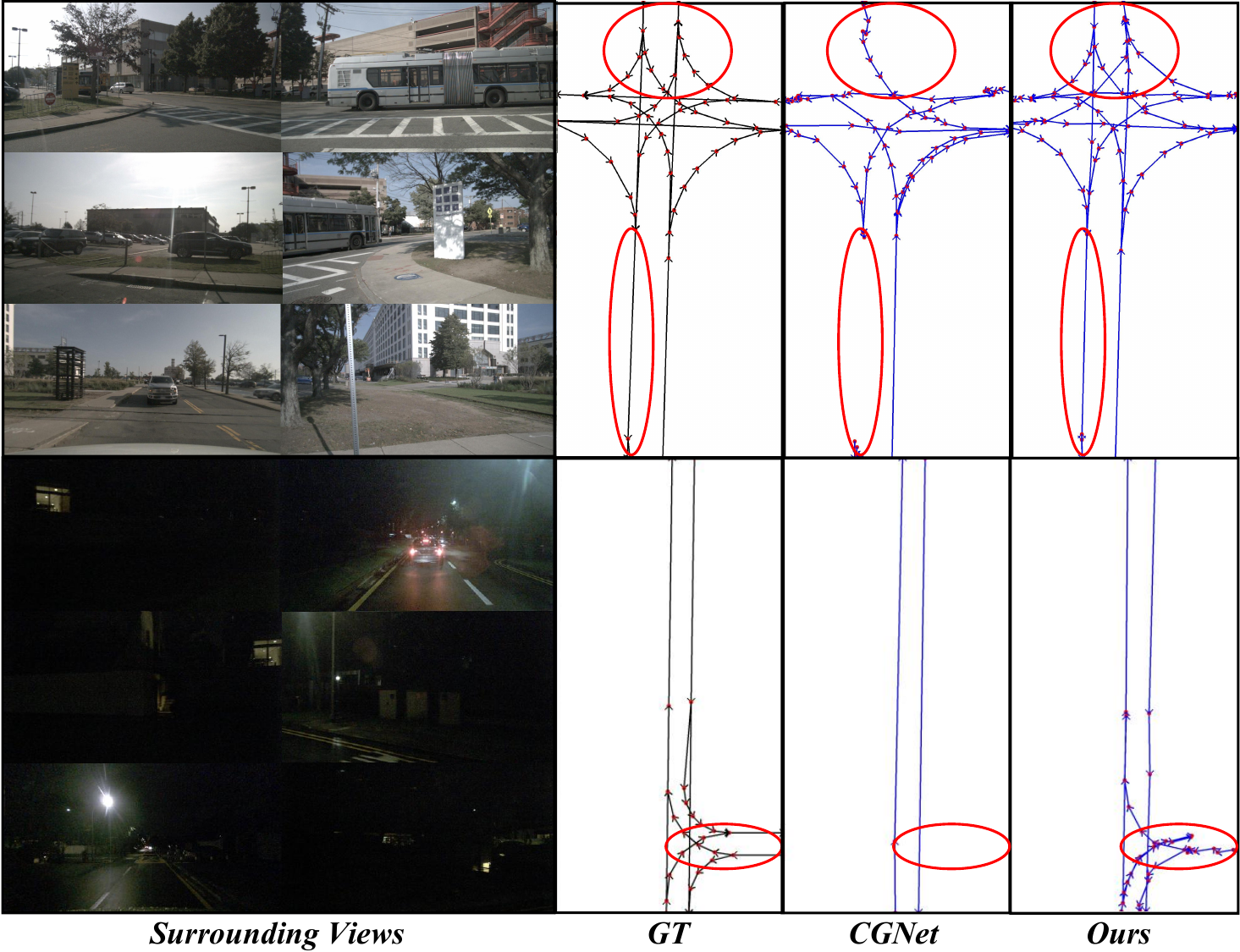}
    \caption{\textbf{The Motivation.} This illustration presents two challenging cases where the state-of-the-art deterministic approach, CGNet, encounters difficulties in handling occlusions or ambiguous visual cues. Our generative framework LaneDiffusion offers a supplementary solution that specifically mitigates these challenges through probabilistic modeling.}
    \label{fig:motivation}
\end{figure}

\section{Introduction}
The topology of drivable lanes, typically represented as centerline graphs, is critical for autonomous driving, as it provides precise path information essential for safe and efficient planning and control of self-driving vehicles in complex driving scenarios \cite{jiang2023vad,openlane}. However, the learning of centerline graphs remains challenging, as it requires highly accurate detection of invisible, logical lane centerlines and robust reasoning about their topological connections.

Most previous centerline graph learning methods are deterministic \cite{vectormapnet,maptr,liu2024mgmap}, whereas generative approaches have received little attention. Despite significant progress, deterministic methods still exhibit inherent limitations. For instance, the state-of-the-art deterministic approach CGNet \cite{CGNet} employs three effective modules to improve lane continuity from different perspectives. However, as illustrated in Fig.~\ref{fig:motivation}, it struggles to capture the full complexity and dynamics of real-world driving in some complex scenarios involving occlusions or ambiguous visual cues. In contrast, the probabilistic modeling capability of generative approaches offers an intuitive supplementary solution to better handle this visual ambiguity.

Despite their potential, employing generative approaches for centerline graph learning presents a non-trivial task. While PolyDiffuse \cite{chen2024polydiffuse} represents a pioneering step in employing generative models for road marker detection—identifying lane dividers, road boundaries, and pedestrian crossings—it falls short of learning logical lane centerlines and inferring lane topology. Furthermore, PolyDiffuse's direct generation of vectorized final results via a diffusion module introduces significant complexities. Specifically, it still struggles to predetermine the number of vectorized outputs and relies heavily on a pre-trained external model as initial prior knowledge. These limitations underscore the need for a more effective generative paradigm for robust centerline graph learning.

To this end, we introduce \textbf{LaneDiffusion}, a novel generative framework for lane centerline learning.  Instead of directly generating vectorized outputs, LaneDiffusion takes a different approach: it generates Bird's Eye View (BEV) features infused with centerline priors. These prior-injected BEV features are then processed by a lane decoder to produce high-quality vectorized centerlines. By operating at the BEV feature level, LaneDiffusion simplifies the learning process, effectively framing it as an image restoration task, and enhances flexibility for subsequent refinement. The LaneDiffusion pipeline unfolds in two key stages: (i) lane prior injection into the BEV feature to construct the diffusion target, and (ii) diffusion module training to generate the prior-injected BEV feature conditioned on the original input feature. This modular design renders LaneDiffusion a versatile and easily integrated add-on for BEV feature-based architectures. Accordingly, we introduce a \textit{Lane Prior Injection Module (LPIM)} to embed lane ground truth (GT) information into the BEV feature, creating the diffusion target.  Subsequently, a \textit{Lane Prior Diffusion Module (LPDM)} learns to model this prior-injected BEV feature. To enhance inference efficiency, we employ a Markov chain that transitions from the prior-injected BEV feature back to the original BEV feature, rather than to Gaussian white noise, significantly reducing the required sampling steps.  The generated lane prior undergoes refinement before being fed into lane decoders, which extract high-quality outputs, including vectorized centerline segments and their connectivity relationships, forming the centerline graph. The core contributions of this work are summarized as follows:
\begin{itemize}
    \item We introduce an end-to-end generative framework \textbf{LaneDiffusion} for centerline graph learning, which pioneers the use of diffusion models to generate lane centerline priors at the BEV feature level, instead of directly predicting vectorized centerlines.
    \item We propose the \textit{Lane Prior Injection Module (LPIM)} for injecting lane priors into BEV features as the diffusion target, and the \textit{Lane Prior Diffusion Module (LPDM)} to model the prior-injected BEV feature.
    \item Extensive experiments on the nuScenes \cite{nuscenes} and Argoverse2 \cite{av2} datasets validate the effectiveness of LaneDiffusion and its components. The significant improvement of performance on both the fine-grained point-level metrics and the segment-level metrics establishes state-of-the-art performance in centerline graph learning and underscores the potential of generative models for handling the centerline graph learning task.
\end{itemize}

\section{Related Works}

\subsection{Lane Graph Learning}
Early lane graph extraction works focused on extracting lane graphs from aerial and satellite imagery \cite{bastani2018roadtracer,li2019topological,he2020sat2graph,he2022lane}. However, the low resolution and frequent occlusions of these images limited these methods to capturing only coarse road structures, rendering them unsuitable for high-precision autonomous driving. Later, methods such as STSU \cite{STSU} introduced online lane graph construction using centerline segments from a single onboard camera \cite{can2022topology,he2022repainting,can2023improving,xia2024dumapnet,xia2025ldmapnet}, but these were constrained by their limited field of view. More recently, TopoNet \cite{TopoNet} utilized a graph-based architecture to predict connectivity among centerline segments using a surround-view camera system. LaneGAP \cite{liao2024laneGAP} proposed a strategy to link centerline segments into complete paths by integrating path detection and vertex merging to preserve continuity. CGNet \cite{CGNet} addressed spatial discontinuities by segmenting lanes into non-overlapping parts and introducing dedicated modules for continuity enhancement. Other recent approaches, such as RNTR \cite{RNTR}, LaneGraph2Seq \cite{peng2024lanegraph2seq} and Topo2Seq \cite{yang2025topo2seq}, leverage auto-regressive sequence-to-sequence modeling to advance lane topology learning. In contrast, our method, LaneDiffusion, is the first to use diffusion models for generating lane centerline priors directly at the BEV feature level, offering a seamless and novel approach to enhance lane graph learning in BEV-based frameworks.

\begin{figure*}[!ht]
    \centering
    \includegraphics[width=\linewidth]{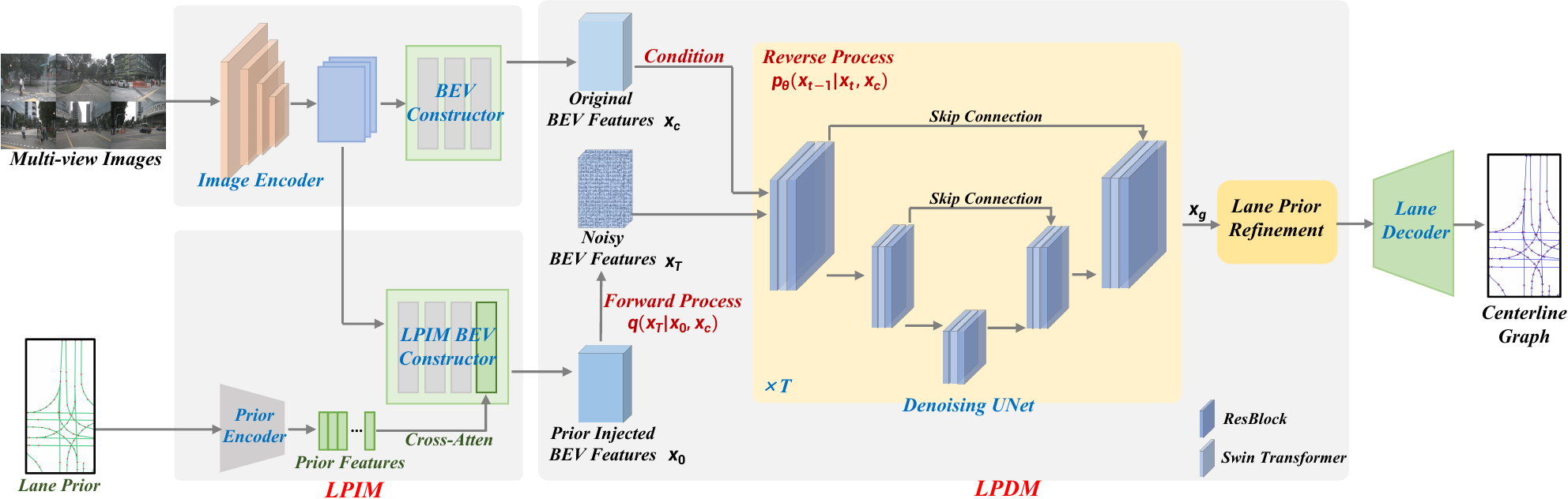}
    \caption{The overall framework of LaneDiffusion. LaneDiffusion comprises two key components: (i) a Lane Prior Injection Module (LPIM), which injects lane priors into BEV features to construct the diffusion target, and (ii) a Lane Prior Diffusion Module (LPDM), which models the prior-injected BEV feature via a diffusion process conditioned on the original feature. The modular design of LaneDiffusion makes it a flexible add-on for BEV feature-based architectures.}
    \label{fig:model}
\end{figure*}

\subsection{Prior Information}
Many studies have explored injecting prior information to improve performance across various autonomous driving tasks. StreamMapNet \cite{streammapnet} incorporates temporal information by propagating queries and fusing BEV features to improve consistency over time. In addition to temporal cues, several works \cite{hu2023collaboration, han2023collaborative, li2024multiagent} leverage spatial vehicle-to-vehicle collaborative perception to achieve better accuracy and a more comprehensive understanding of the environment. Topo2D \cite{li2024enhancing} enhances 3D lane perception by initializing 3D queries and positional embeddings with 2D lane instances, while SatForHDMap \cite{zhang2024enhancing} uses satellite imagery to build high-precision maps in real-time. Similarly, HRMapNet \cite{gao2024complementing} integrates historical rasterized maps to refine online vectorized map perception. Other recent studies \cite{ort2022maplite, jiang2024p, SMERF} investigate the use of standard-definition (SD) maps for autonomous driving. While SD maps provide structured road topology, their inherent inaccuracies limit their effectiveness. Jia et al. \cite{jia2024crowdsourcing} utilize trajectory data to enhance lane segmentation and topology reasoning. While these methods offer benefits, they often introduce system complexity, pose alignment challenges and rely on external data that may not be available during inference. By leveraging diffusion models, LaneDiffusion operates solely on image data while injecting prior knowledge, ensuring a self-contained and robust solution for lane graph learning.

\subsection{Diffusion Models}
Denoising Diffusion Probabilistic Models (DDPMs) \cite{ddpm} represent a powerful generative model grounded in Markov chains. They have been widely applied \cite{huang2025dreamfuse,huang2025dreamlayer,lou2024multi,he2024wildvidfit,du2024highlighted}, especially in tasks such as image synthesis \cite{stablediffusion}, semantic segmentation \cite{amit2021segdiff} and object detection \cite{diffusiondet}, and have gained increasing interest in autonomous driving. For example, MotionDiffuser \cite{jiang2023motiondiffuser} applies diffusion models to multi-agent motion prediction. MagicDrive \cite{gao2023magicdrive} generates street views under geometric constraints, while the DriveDreamer series \cite{wang2024drivedreamer, zhao2024drivedreamer2, zhao2024drivedreamer4d} uses diffusion models for 4D scene reconstruction. DiffMap \cite{jia2024diffmap} proposes a generative framework for map segmentation, which is not able to generate vectorized results. PolyDiffuse \cite{chen2024polydiffuse} uses DDPMs to generate structured vectorized map elements. Although PolyDiffuse shows the potential of DDPMs in reconstruction tasks, it relies on pre-trained external models for initialization and struggles with missing geometry in the initial reconstruction. In contrast, LaneDiffusion eliminates the need for external initialization and pioneers to detect logical lane centerlines and infer their topological connections via a BEV-level centerline prior diffusion generation paradigm.

\section{Methodology}
In this section, we first introduce the definition of the centerline graph learning task and present the formulations of two types of centerline graphs in Sec.~\ref{sec:task-def}. Next, we provide an overview of our proposed framework in Sec.~\ref{sec:overview}. The details of the Lane Prior Injection Module (LPIM) and Lane Prior Diffusion Module (LPDM) are described in Sec.~\ref{sec:lpim} and Sec.~\ref{sec:lpdm}, respectively. Finally, the optimization method for the framework is introduced in Sec.~\ref{sec:optimization}.

\subsection{Task Definition}
\label{sec:task-def}
Given images captured by a vehicle's surround-view cameras, the task of centerline graph learning is to detect non-overlapping centerline segments in the Bird’s Eye View (BEV) and infer the topology among these segments to form a centerline graph. Based on the granularity required by different metrics (detailed in Sec.~\ref{sec:metrics}), both a segment-level graph and a fine-grained point-level graph are formulated. Specifically, the segment-level graph is defined as $G=(V, E)$, where the vertices $V = \{P_i\}_{i=1}^{|V|}$ represent the set of centerline segments, and $|V|$ denotes the total number of segments. The edges $E \subseteq V \times V$ denote the connectivity among segments. Each segment is represented as an ordered set of points, denoted by $P = [p_1, p_2, \ldots, p_N]$, with each point $p = (x, y) \in \mathbb{R}^2$. The connectivity is encoded in the adjacency matrix of the graph $G$. In contrast, the fine-grained point-level graph $\ddot{G} = (\ddot{V}, \ddot{E})$ is constructed such that $\ddot{V}$ comprises all points of the polylines, and $\ddot{E}$ represents the connectivity among these points.
\subsection{Framework Overview}
\label{sec:overview}
The overall framework of LaneDiffusion is illustrated in Fig.~\ref{fig:model}. As highlighted in the Introduction, it can be seamlessly integrated as an add-on to BEV feature-based architectures. To clearly demonstrate the performance improvement brought by LaneDiffusion, we adopt the state-of-the-art CGNet \cite{CGNet} as our primary baseline, which extracts multi-view features with an image encoder and then transforms these features into the BEV space with a BEV constructor. The proposed LaneDiffusion consists of two key steps: (i) \textit{BEV-level lane prior injection} for constructing the diffusion target and (ii) \textit{BEV-level lane prior modeling} via a diffusion module to enhance lane centerline graph learning. In step (i), we introduce the \textit{Lane Prior Injection Module (LPIM)}, which injects lane prior into the BEV feature in a SMERF-like \cite{SMERF} manner. In step (ii), an efficient \textit{Lane Prior Diffusion Module (LPDM)} is employed to model the prior-injected BEV feature with a relatively low sampling step budget. The generated BEV feature is further refined through a lane prior refinement mechanism before being passed to the lane decoder to extract high-quality results, including the prediction of vectorized centerline segments and their corresponding connectivity relationships represented by the segment-level and point-level graphs.


\subsection{Lane Prior Injection Module (LPIM)}
\label{sec:lpim}
To establish robust BEV feature-level diffusion targets for optimizing the subsequent diffusion module, we employ a \textit{Lane Prior Injection Module (LPIM)} that embeds prior knowledge directly into the BEV feature. The design of LPIM primarily focuses on the selection, encoding and injection of prior knowledge.

\paragraph{Prior Knowledge Selection.}
A crucial aspect of LPIM is selecting the appropriate prior knowledge to integrate. Existing works involving the injection of external prior knowledge often rely on secondary sources, such as SDMap \cite{SMERF} or crowdsourced trajectories \cite{jia2024crowdsourcing}, rather than the most precise information. This is mainly due to concerns over the availability and acquisition cost of such priors during inference. However, LPIM employs external priors solely to construct the diffusion target, eliminating the need for them during inference and thus avoiding issues related to acquisition feasibility or potential ground truth (GT) leakage. Therefore, we directly utilize the highest-precision \textit{lane centerline GT} as the prior, which specifically denotes the ground truth centerlines of lanes in the BEV space.

\paragraph{Prior Knowledge Encoding.}
Given the lane centerline GTs for each sample, a Prior Encoder is proposed for prior knowledge encoding. To be specific, each of the $M$ centerlines is evenly sampled for $N$ fixed points, denoted as $\{(x_{i}, y_{i})\}_{i=1}^{N}$. Sinusoidal embeddings with varying frequencies are used to embed the point locations, which enhance the sensitivity to positional variations \cite{vaswani2017attention}. Following SMERF, the embedded lane priors are further encoded using a transformer encoder consisting of $L$ multi-head self-attention layers to capture global geometric and semantic information. The output has a shape of $M \times H$, where $H$ is the feature dimension produced by the self-attention layers.

\paragraph{Prior Knowledge Injection.}
Drawing inspiration from SMERF, we inject the encoded prior knowledge in the BEV features via a modified version of the CGNet BEV constructor. To be specific, the encoded $M \times H$ lane prior features are cross-attended with the intermediate BEV feature representations after each spatial cross-attention operation in the BEV constructor. This modified BEV constructor is termed as the LPIM BEV Constructor.


\subsection{Lane Prior Diffusion Module (LPDM)}
\label{sec:lpdm}
With solid diffusion targets generated by LPIM, we introduce the \textit{Lane Prior Diffusion Module (LPDM)}, which employs Denoising Diffusion Probabilistic Models (DDPMs) to model BEV features embedded with prior knowledge. In this section, we first provide a brief overview of DDPMs, followed by a detailed description of the proposed LPDM.

\paragraph{Preliminary: DDPMs.}
DDPMs are generative models that approximate data distributions by introducing noise in a forward process and denoising in a reverse process. In the forward process, a $T$-step Markov chain transforms clean data $\mathbf{x}_{0} \sim q(\mathbf{x}_{0})$ into a specified prior distribution (typically standard Gaussian):
\begin{equation}
    q(\mathbf{x}_t|\mathbf{x}_{t-1}) = \mathcal{N}(x_{t}; \sqrt{1-\beta_{t}} \mathbf{x}_{t-1}, \beta_{t} \mathbf{I}),
\end{equation}
where $t = 1, \dots, T$ and $\mathbf{x}_{t}$ denotes the latent variable at step $t$. The sequence $\{\beta_{t}\}_{t=1}^{T}$ controls the noise variance at each step, while $\mathcal{N}$ represents a Gaussian distribution and $\mathbf{I}$ is the identity matrix. The forward process can be expressed in closed form for any step $t$ as:
\begin{equation}
    \label{eq:inte}
    q(\mathbf{x}_t |\mathbf{x}_{0}) = \mathcal{N}(x_{t}; \sqrt{\overline{\alpha}_{t}} \mathbf{x}_{0}, (1-\overline{\alpha}_{t})\mathbf{I}),
\end{equation}
where $\overline{\alpha}_{0} = 1$, $\overline{\alpha}_{t} = \overline{\alpha}_{t-1}\alpha_{t}$ and $\alpha_{t} = 1 - \beta_{t}$. Thus, $\mathbf{x}_{t}$ can be directly sampled as $\mathbf{x}_{t} = \sqrt{\overline{\alpha}_{t}}\mathbf{x}_{0} + \sqrt{1-\overline{\alpha}_{t}} \boldsymbol{\epsilon}$, with $\boldsymbol{\epsilon} \sim \mathcal{N}(0, \mathbf{I})$. In the reverse process, the clean data is generated by progressively denoising a sample $\mathbf{x}_{t}$ using a learnable denoising network, following:
\begin{equation}
    p_{\boldsymbol{\theta}}(\mathbf{x}_{t-1}|\mathbf{x}_{t}) = \mathcal{N}(\mathbf{x}_{t-1}; \boldsymbol{\mu}_{\boldsymbol{\theta}}(\mathbf{x}_{t}, t), \boldsymbol{\Sigma}_{\boldsymbol{\theta}}(\mathbf{x}_{t}, t)),
\end{equation}
where $\boldsymbol{\mu}_{\boldsymbol{\theta}}$ and $\boldsymbol{\Sigma}_{\boldsymbol{\theta}}$ represent the mean and covariance predicted by the network, and $\boldsymbol{\theta}$ are the learnable network parameters.

\begin{algorithm}
    \SetAlgoLined 
	\caption{Training of LPDM}
    \label{alg:training}
	\KwIn{Diffusion target set $\boldsymbol{\mathcal{X}_{0}}$, Condition feature set $\boldsymbol{\mathcal{X}_{\textrm{cond}}}$, Diffusion step number $T$, Maximum iteration number $max\_iter$}
    $cur\_iter := 1$\;
	\While{cur\_iter $\leq$ max\_iter}{
        $\mathbf{x}_0 \sim \boldsymbol{\mathcal{X}_{0}}$, $\mathbf{x}_{\textrm{c}} \sim \boldsymbol{\mathcal{X}_{\textrm{cond}}}$\;
		$t$ $\sim$ $Uniform$ $(\{1, 2, \dots, T\})$\;
        $\mathbf{x}_t$ $\sim$ $q(\mathbf{x}_t|\mathbf{x}_0, \mathbf{x}_{\textrm{c}})$ (see Eq.~\ref{eq:lsdm1})\;
        Take gradient descent step on $\nabla \mathcal{L}_{\textrm{diff}}$ (see Eq.\ref{eq:diffloss})\;
        $cur\_iter$ := $cur\_iter$ + 1\;
	}
\end{algorithm}
\vspace{-0.6cm}
\begin{algorithm}
    \SetAlgoLined 
	\caption{Sampling of LPDM}
    \label{alg:sampling}
	\KwIn{Condition feature $\mathbf{x}_{\textrm{c}}$, Diffusion step number $T$}
	\KwOut{Denoised feature $\mathbf{x}_{\textrm{g}}$}
    $\mathbf{x}_T \sim \mathcal{N}(\mathbf{x}_T; \mathbf{x}_{\textrm{c}}, \kappa^2 \eta_T \mathbf{I})$\;
	\For{$t := T, T-1, \dots, 1$}{
        $\boldsymbol{\epsilon} \sim \mathcal{N}(\boldsymbol{\epsilon}; \mathbf{0}, \mathbf{I})$ if $t > 1$ else $\boldsymbol{\epsilon} := 0$\;
        $\boldsymbol{\mu} := \frac{\eta_{t-1}}{\eta_t} \mathbf{x}_{t} + \frac{\gamma_{t}}{\eta_{t}} f_{\boldsymbol{\theta}}(\mathbf{x}_{t}, \mathbf{x}_{\textrm{c}}, t)$ (see Eq.~\ref{eq:lsdm4})\;
        $\mathbf{x}_{t-1} := \boldsymbol{\mu} + \kappa \sqrt{\frac{\eta_{t-1} \gamma_t}{eta_t}} \boldsymbol{\epsilon}$\;
	}
    $\mathbf{x}_{\textrm{g}} := \mathbf{x}_{0}$\;
    return $\mathbf{x}_{\textrm{g}}$
\end{algorithm}
\vspace{-0.4cm}

\begin{table*}[!ht]
    \centering
    \begin{tabular}{lc|ccccc}
    \toprule
    Method & Epoch & GEO F1 $\uparrow$ & TOPO F1 $\uparrow$ & JTOPO F1 $\uparrow$ & APLS $\uparrow$ & SDA $\uparrow$ \\
    \midrule
    STSU \cite{STSU} & 200 & 33.0 & 20.6 & 13.9 & 11.0 & 7.0 \\
    HDMapNet \cite{hdmapnet} & 30 & 45.5 & 20.0 & 14.8 & 25.9 & 0.5 \\
    VectorMapNet \cite{vectormapnet} & 24 & 48.4 & 38.1 & 27.9 & 10.3 & 7.2 \\
    TopoNet \cite{TopoNet} & 24 & 50.8 & 39.6 & 31.9 & 26.2 & 5.6 \\
    MapTR \cite{maptr} & 24 & 53.3 & 39.7 & 32.4 & 26.8 & 7.6 \\
    CGNet \cite{CGNet} & 24 & 54.7 & 42.2 & 34.1 & 30.7 & 8.8 \\
    \midrule
    LaneDiffusion (Ours) & 24 & \textbf{58.9 (+4.2)} & \textbf{46.8 (+4.6)} & \textbf{38.8 (+4.7)} & \textbf{37.1 (+6.4)} & \textbf{10.6 (+1.8)} \\
    \bottomrule
    \end{tabular}
    \caption{Comparison with SOTAs on nuScenes using point-level metrics.}
    \label{tab:nuScenes-point}
\end{table*}
\begin{table*}[!ht]
    \centering
    \begin{tabular}{lc|cccc}
    \toprule
    Method & Epoch & IoU $\uparrow$ & mAP$_{cf}$ $\uparrow$ & DET$_{l}$ $\uparrow$ & TOP$_{ll}$ $\uparrow$  \\
    \midrule
    VectorMapNet \cite{vectormapnet} & 24 & 29.8 & 25.5 & 16.1 & 0.5 \\
    TopoNet \cite{TopoNet} & 24 & 51.3 & 31.6 & 17.9 & 0.8 \\
    MapTR \cite{maptr} & 24 & 53.7 & 33.1 & 18.9 & 1.0 \\
    CGNet \cite{CGNet} & 24 & 56.3 & 35.2 & 22.0 & 1.3 \\
    \midrule
    LaneDiffusion (Ours) & 24 & \textbf{58.6 (+2.3)} & \textbf{41.6 (+6.4)} & \textbf{28.8 (+6.8)} & \textbf{3.4 (+2.1)} \\
    \bottomrule
    \end{tabular}
    \caption{Comparison with SOTAs on nuScenes using segment-level metrics.}
    \label{tab:nuScenes-segment}
    \vspace{-0.4cm}
\end{table*}

\paragraph{LPDM.}
While the standard DDPM paradigm is effective for general tasks like image generation, it often requires many (typically hundreds or even thousands) sampling steps when starting from a standard Gaussian prior. In contrast, LPDM initializes the reverse process from a prior based on the original BEV feature, as inspired by ResShift \cite{yue2023resshift}. This reduces the required diffusion steps and improves efficiency. In the \textbf{forward process}, we transition from $\mathbf{x}_{0}$ to $\mathbf{x}_{\textrm{c}}$ by gradually shifting their residual $\mathbf{x}_{\textrm{res}} = \mathbf{x}_{\textrm{c}} - \mathbf{x}_{0}$ through a Markov chain of length $T$. The marginal transition distribution at each timestep $t$ is:
\begin{equation}
    \label{eq:lsdm1}
    q(\mathbf{x}_t|\mathbf{x}_{0}, \mathbf{x}_{c}) = \mathcal{N}(x_{t}; \mathbf{x}_{0}+\eta_{t}\mathbf{x}_{\textrm{res}}, \kappa^2 \eta_{t} \mathbf{I}),
\end{equation}
where hyper-parameter $\kappa$ controls the noise variance, while sequence $\{\eta_t\}_{t=1}^{T}$ is the shifting schedule increasing monotonically with $t$ from $\eta_1 \rightarrow 0$ to $\eta_T \rightarrow 1$, which is implemented as a non-uniform geometric schedule following:
\begin{equation}
    \sqrt{\eta_t} = \begin{cases}
        \textrm{min}\left(\frac{0.04}{\kappa}, \sqrt{0.001}\right), \quad t = 1,\\
        \sqrt{\eta_1} \times b_0^{\zeta_t}, \quad t = 2, 3, \dots, T-1,\\
        \sqrt{0.999}, \quad t = T,
    \end{cases}
\end{equation}
where
\begin{equation}
    \zeta_t = \left(\frac{t-1}{T-1}\right)^p \times (T-1), \quad b_0 = e^{\frac{1}{2(T-1)}\textrm{log}\frac{\eta_T}{\eta_1}},
\end{equation}
where $p$ is the hyper-parameter that controls the growth rate of $\sqrt{\eta_t}$. In the \textbf{reverse process}, LPDM denoises $\mathbf{x}_{t}$ using the original BEV feature $\mathbf{x}_{\textrm{c}}$ as a condition, following:
\begin{equation}
    \label{eq:lsdm2}
    p_{\boldsymbol{\theta}}(\mathbf{x}_{t-1}|\mathbf{x}_{t}, \mathbf{x}_{\textrm{c}}) = \mathcal{N}(\mathbf{x}_{t-1}; \boldsymbol{\mu}_{\boldsymbol{\theta}}(\mathbf{x}_{t}, \mathbf{x}_{\textrm{c}}, t), \boldsymbol{\Sigma}_{\boldsymbol{\theta}}(\mathbf{x}_{t}, \mathbf{x}_{\textrm{c}}, t)).
\end{equation}
LPDM uses a swin transformer \cite{liu2021swin} based UNet $f_{\boldsymbol{\theta}}$ as the denoising network with optimization achieved by minimizing the negative evidence lower bound \cite{sohl2015deep, ddpm} with respect to the target distribution $q(\mathbf{x}_{t-1}|\mathbf{x}_{t}, \mathbf{x}_{0}, \mathbf{x}_{\textrm{c}})$:
\begin{equation}
    \label{eq:lsdm5}
    \boldsymbol{\theta}^{*}  = \textrm{argmin}_{\boldsymbol{\theta}} \sum_{t} D_{\textrm{KL}} (q(\mathbf{x}_{t-1}|\mathbf{x}_{t}, \mathbf{x}_{0}, \mathbf{x}_{\textrm{c}}) || p_{\boldsymbol{\theta}}(\mathbf{x}_{t-1}|\mathbf{x}_{t}, \mathbf{x}_{\textrm{c}})),
\end{equation}
where $D_{\textrm{KL}}(\cdot||\cdot)$ is the Kullback-Leibler (KL) divergence, while $\boldsymbol{\theta}^{*}$ denotes the optimized parameters. According to Eq.~\ref{eq:lsdm1}, the targeted distribution can be rendered tractable and given by:
\begin{equation}
    \label{eq:lsdm3}
    q(\mathbf{x}_{t-1}|\mathbf{x}_{t}, \mathbf{x}_{0}, \mathbf{x}_{\textrm{c}}) = \mathcal{N}\left(\mathbf{x}_{t-1}; \frac{\eta_{t-1}}{\eta_t} \mathbf{x}_{t} + \frac{\gamma_{t}}{\eta_{t}}\mathbf{x}_{0}, \kappa^2\frac{\eta_{t-1}}{\eta_t}\gamma_{t}\mathbf{I} \right),
\end{equation}
where $\gamma_{1} = \eta_{1}$ and $\gamma_{t} = \eta_{t} - \eta_{t-1}$ for $t > 1$. Please refer to Sec.~A of the supplementary materials for details about the calculation of this derivation. It can be observed from Eq.~\ref{eq:lsdm3} that the variance parameter is independent of $\mathbf{x}_{t}$ and $\mathbf{x}_{\textrm{c}}$, so the variance item in Eq.~\ref{eq:lsdm2} can be set as $\boldsymbol{\Sigma}_{\boldsymbol{\theta}}(\mathbf{x}_{t}, \mathbf{c}, t) = \kappa^2\frac{\eta_{t-1}}{\eta_t}\gamma_{t}\mathbf{I}$, while the mean item is re-parameterized as:
\begin{equation}
    \label{eq:lsdm4}
    \boldsymbol{\mu}_{\boldsymbol{\theta}}(\mathbf{x}_{t}, \mathbf{x}_{\textrm{c}}, t) = \frac{\eta_{t-1}}{\eta_t} \mathbf{x}_{t} + \frac{\gamma_{t}}{\eta_{t}} f_{\boldsymbol{\theta}}(\mathbf{x}_{t}, \mathbf{x}_{\textrm{c}}, t).
\end{equation}
By comparing Eq.~\ref{eq:lsdm3} and Eq.~\ref{eq:lsdm4}, the goal of the denoising UNet $f_{\boldsymbol{\theta}}$ is now transformed to predict the diffusion target $\mathbf{x}_{0}$. The objective function in Eq.~\ref{eq:lsdm5} can be simplified as:
\begin{equation}
    \label{eq:diffloss}
    \boldsymbol{\theta}^{*} = \textrm{argmin}_{\boldsymbol{\theta}} \mathcal{L}_{\textrm{diff}} = \textrm{argmin}_{\boldsymbol{\theta}} w_t ||f_{\boldsymbol{\theta}}(\mathbf{x}_{t}, \mathbf{x}_{\textrm{c}}, t) - \mathbf{x}_0||_2^2,
\end{equation}
where $w_t = \frac{\alpha_2}{2\kappa^2 \eta_t \eta_{t-1}}$. This design improves the training stability. Given a noise map sampled from the prior distribution based on $\mathbf{x}_{\textrm{c}}$, the optimized LPDM generates the lane prior generated BEV feature $\mathbf{x}_{\textrm{g}}$ after $T$ denoising steps. After obtaining the prior-generated BEV feature $\mathbf{x}_{\textrm{g}}$, a lane prior refinement mechanism is applied to further integrate it with the original feature, thereby preserving and enhancing the structural knowledge already captured, which is formulated as:
\begin{equation}
    \hat{\mathbf{x}}_{\textrm{g}} = \textrm{LPR}([\mathbf{x}_{\textrm{g}}, \mathbf{x}_{\textrm{c}}]),
\end{equation}
where $[\cdot, \cdot]$ represents the concatenation of features, and $\textrm{LPR}(\cdot)$ denotes the lane prior refinement operation, which is implemented as an encoder-decoder structure (please refer to Sec.~B of the supplementary materials for more details). The resulting enhanced feature $\hat{\mathbf{x}}_{\textrm{g}}$ is then passed to lane decoders for vectorized result extraction, similar to standard BEV features, ensuring that LaneDiffusion integrates seamlessly with BEV-based architectures.

\subsection{Optimization}
\label{sec:optimization}
The optimization of LaneDiffusion consists of three stages. In \textbf{stage-I}, we train LPIM using the loss constraints (termed as $\mathcal{L}_{\textrm{lane}}$) proposed by the CGNet baseline, formulated as follows:
\begin{equation}
\label{eq:loss-base}
\begin{aligned}
    \mathcal{L}_{\textrm{lane}} = \lambda_1 \mathcal{L}_{\textrm{cls}} + \lambda_2 \mathcal{L}_{\textrm{poly}} + \lambda_3 \mathcal{L}_{\textrm{topo}} \\ + \lambda_4 \mathcal{L}_{\textrm{dir}} + \lambda_5 \mathcal{L}_{\textrm{bezier}} + \lambda_6 \mathcal{L}_{\textrm{ja}},
\end{aligned}
\end{equation}
where the six subitems denote the loss constraints on classification, polyline point distance, topology, lane direction, Bézier space consistency and junction points, respectively. In \textbf{stage-II}, the optimized LPIM is frozen to provide diffusion targets for the training of LPDM, subject to the constraint $\mathcal{L}_{\textrm{diff}}$ as defined in Eq.~\ref{eq:diffloss}. The training algorithm of training LPDM is summarized in Alg.~\ref{alg:training}. In \textbf{stage-III}, LPDM is fixed and switched to the sampling mode (summarized in Alg.~\ref{alg:sampling}) and a lane decoder same as the baseline model is optimized using $\mathcal{L}_{\textrm{lane}}$.


\begin{figure*}[!ht]
    \centering
    \includegraphics[width=\linewidth]{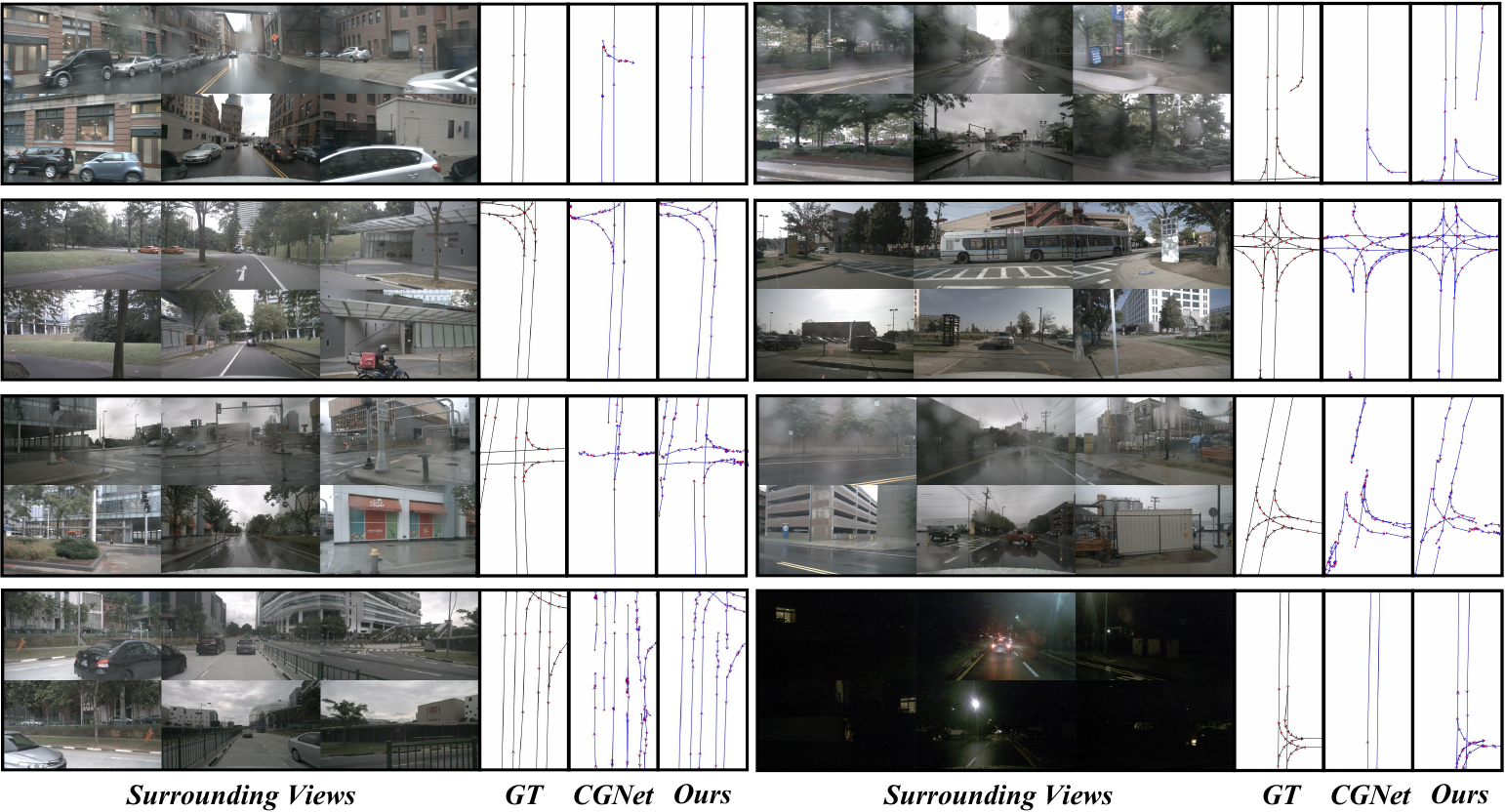}
    \vspace{-0.4cm}
    \caption{Qualitative comparisons under different weather and lighting conditions on nuScenes.}
    \label{fig:visualization}
    \vspace{-0.4cm}
\end{figure*}

\section{Experiments}
\subsection{Dataset and Metric}
\paragraph{Dataset.}
The proposed method is evaluated on the challenging nuScenes \cite{nuscenes} and Argoverse2 \cite{av2} datasets. The nuScenes dataset consists of 1,000 driving scenes, each lasting approximately 20 seconds, with key-frame annotations provided at a frequency of 2 Hz. Each sample offers a complete 360° horizontal field of view (FOV) around the ego-vehicle. In contrast, the Argoverse2 dataset includes 1,000 driving logs, each offering 15 seconds of 20 Hz RGB image sequences from 7 cameras, along with a log-level 3D vectorized map. Notably, both the nuScenes and Argoverse2 datasets provide direct access to centerline segments along with their topological connectivity, making them ideal for map-based research and development in autonomous driving.

\paragraph{Evaluation Metrics.}
\label{sec:metrics}
To ensure a fair comparison with previous methods, we follow the evaluation protocol of CGNet \cite{CGNet} and employ both fine-grained point-level metrics (GEO F1 \cite{he2022lane}, TOPO F1 \cite{he2022lane}, JTOPO F1 \cite{liao2024laneGAP}, APLS \cite{van2018spacenet}, SDA \cite{buchner2023learning}) and segment-level metrics (IoU \cite{hdmapnet}, mAP$_{cf}$ \cite{vectormapnet}, DET$_{l}$ \cite{openlane}, TOP$_{ll}$ \cite{openlane}) to evaluate LaneDiffusion. As described in Sec.~\ref{sec:task-def}, both a segment-level graph $G=(V, E)$ and a fine-grained point-level graph $\ddot{G}=(\ddot{V},\ddot{E})$ are formulated based on the granularity required by different metrics. For further details on the definition of each metric, please refer to Sec.~C of the supplementary materials.

\subsection{Implementation Details}
\label{sec:imple}
Model training and baseline reproduction were conducted using PyTorch on 8 Tesla V100 GPUs. First, LPIM was optimized for 24 epochs with the AdamW optimizer and a cosine annealing learning rate schedule, with an initial learning rate of $6 \times 10^{-4}$. Next, LPDM was trained from scratch for 24 epochs using the AdamW optimizer and an initial learning rate of $6 \times 10^{-5}$. Finally, a lane decoder same as the baseline model is trained from scratch with the AdamW optimizer (initial learning rate of $6 \times 10^{-4}$) while keeping LPDM fixed for 24 epochs. The weighting factors $\{\lambda_1, \lambda_2, \lambda_3, \lambda_4, \lambda_5, \lambda_6\}$ in Eq.~\ref{eq:loss-base} are set to $\{2, 5, 1, 0.005, 0.01, 0.1\}$, respectively. The batch size on each GPU is set to 2. The perception range was set to [-15.0 m, 15.0 m] along the X-axis and [-30.0 m, 30.0 m] along the Y-axis. We reproduced the state-of-the-art CGNet \cite{CGNet} as our primary baseline, which employs ResNet-50 \cite{resnet} with FPN \cite{fpn} as the backbone and uses GKT \cite{GKT} to extract BEV features $\in \mathbb{R}^{256 \times 200 \times 100}$, while deformable attention \cite{deformable} enables query interactions with BEV features in the centerline decoder. In LPIM, the number of centerline priors $M$ per sample is kept flexible, whereas the number of points per polyline is fixed at $N=20$. The feature dimension $H$ is set to 256. The dimension of positional embeddings is set to $d=32$, while the layer number is set to $L=6$ for the prior transformer encoder. In LPDM, the noise variance hyperparameter is set to $\kappa=2.0$, while the $\sqrt{\eta_t}$ growth rate controlling hyperparameter is set to $p=0.3$. The diffusion step number is by default set to $T=15$. Furthermore, during inference, in consideration of the stochastic nature of generative models, the LPDM sampling process is executed three times, and the average of the generated features is passed to subsequent modules to enhance stability.

\vspace{-0.4cm}
\subsection{Comparison with State-of-the-Arts}
\paragraph{Quantitative Results.}
We quantitatively compare LaneDiffusion with previous state-of-the-art methods on the nuScenes dataset using point-level and segment-level metrics, as shown in Tab.~\ref{tab:nuScenes-point} and Tab.~\ref{tab:nuScenes-segment}, respectively. Our results on the nuScenes validation set show that LaneDiffusion outperforms all competitors by a significant margin. Specifically, LaneDiffusion exceeds the CGNet baseline by $4.2\%$, $4.6\%$, $4.7\%$, $6.4\%$ and $1.8\%$ on the GEO F1, TOPO F1, JTOPO F1, APLS and SDA metrics, respectively. These improvements demonstrate that leveraging diffusion models to model lane centerline priors at the BEV feature level enhances both point detection and topology prediction. Furthermore, comparisons on the Argoverse2 dataset (Tab.~\ref{tab:av2}) further confirm the effectiveness of our proposed approach.
\vspace{-0.4cm}
\paragraph{Qualitative Analysis.}
We also present qualitative comparisons with the CGNet baseline under various weather and lighting conditions on the nuScenes dataset in Fig.~\ref{fig:visualization}, highlighting the performance improvements achieved by LaneDiffusion. It can be observed that our generative framework LaneDiffusion effectively captures missing lane structures that the deterministic baseline fails to detect in these cases.
\begin{table*}[!ht]
    \centering
    \begin{tabular}{lc|ccccc}
    \toprule
    Method & Epoch & TOPO F1 $\uparrow$ & JTOPO F1 $\uparrow$ & APLS $\uparrow$ & SDA $\uparrow$ & TOP$_{ll}$ $\uparrow$ \\
    \midrule
    TopoNet \cite{TopoNet} & 6 & 30.2 & 23.7 & 15.3 & 7.7 & 0.3 \\
    MapTR \cite{maptr} & 6 & 42.8 & 33.5 & 22.3 & 13.6 & 0.5 \\
    CGNet \cite{CGNet} & 6 & 44.5 & 34.6 & 23.6 & 13.7 & 0.5 \\
    \midrule
    LaneDiffusion (Ours) & 6 & \textbf{47.0(+2.5)} & \textbf{36.4(+1.8)} & \textbf{25.8(+2.2)} & \textbf{14.4(+0.7)} & \textbf{1.8(+1.3)} \\
    \bottomrule
    \end{tabular}
    \vspace{-0.2cm}
    \caption{Comparison with SOTAs on Argovese2 val.}
    \label{tab:av2}
    \vspace{-0.4cm}
\end{table*}
 
\begin{table}[!ht]
    \centering
    \resizebox{\columnwidth}{!}{
    \begin{tabular}{c|c|cccc}
    \toprule
    $T$ & Paradigm & TOPO F1 $\uparrow$ & JTOPO F1 $\uparrow$ & APLS $\uparrow$ & SDA $\uparrow$ \\
    \midrule
    5 & Overall & 45.6 & 37.5 & 35.5 & 9.5 \\
    5 & Sampling & 44.8 & 36.4 & 33.1 & 9.2 \\
    15 & Sampling & \textbf{46.8} & \textbf{38.8} & \textbf{37.1} & \textbf{10.6} \\
    30 & Sampling & 46.0 & 37.6 & 36.3 & 10.2 \\
    \bottomrule
    \end{tabular}
    }
    \vspace{-0.2cm}
    \caption{Ablation analysis for sampling step number $T$ and LPDM training paradigm.}
    \label{tab:abl-T}
    \vspace{-0.2cm}
\end{table}
\begin{table}[!ht]
    \centering
    \resizebox{\columnwidth}{!}{
    \begin{tabular}{c|cccc}
    \toprule
    Method & TOPO F1 $\uparrow$ & JTOPO F1 $\uparrow$ & APLS $\uparrow$ & SDA $\uparrow$ \\
    \midrule
    No Refine. & 43.9 & 35.9 & 33.3 & 9.2 \\
    Concat.\&FC & 45.6 & 37.0 & 35.7 & 9.8 \\
    Concat.\&ED & \textbf{46.8} & \textbf{38.8} & \textbf{37.1} & \textbf{10.6} \\
    Add.\&FC & 45.3 & 37.2 & 35.7 & 10.2 \\
    Add.\&ED & \textbf{46.8} & 38.5 & 36.9 & \textbf{10.6} \\
    \bottomrule
    \end{tabular}
    }
    \vspace{-0.2cm}
    \caption{Ablation analysis for feature refinement mechanism.}
    \label{tab:abl-refine}
    \vspace{-0.4cm}
\end{table}

\subsection{Ablation Analysis}
In this section, we present a series of ablation studies on nuScenes to evaluate the effectiveness of key components and the impact of hyperparameter settings. All experiments are conducted for 24 epochs.
\vspace{-0.4cm}
\paragraph{Diffusion Step Number $T$.}
Adopting the transition distribution defined in Eq.~\ref{eq:lsdm1} enables a substantial reduction in the required diffusion steps $T$ of the Markov chain. Tab.~\ref{tab:abl-T} summarizes the performance variation under different configurations of $T$. It can be observed that $T=15$ yields the best results, with improvements of $2.0\%$, $2.4\%$, $4.0\%$ and $1.4\%$ in the TOPO F1, JTOPO F1, APLS and SDA metrics, respectively. In contrast, increasing $T$ to 30 results in lower inference efficiency and slightly inferior performance.
\vspace{-0.4cm}
\paragraph{Training Paradigm for LPDM.}
We further explore the impact of different training paradigms for LPDM in Tab.~\ref{tab:abl-T}. Here, \textit{Sampling} refers to the commonly used random sampling of time step $t$ in each training iteration, while \textit{Overall} denotes the approach of performing full denoising to generate a clean feature, starting from a sampled noise as in the inference stage, and calculating the loss for intermediate features at each timestep $t$. As shown by the results, the \textit{Overall} paradigm outperforms the \textit{Sampling} paradigm when $T=5$, suggesting that performance may improve with larger values of $T$, too. However, due to GPU memory limitations, we still use the \textit{Sampling} paradigm for larger $T$.
\vspace{-0.5cm}
\paragraph{Lane Prior Refinement Mechanism in LPDM.}
We conduct additional ablation studies to examine the impact of the feature refinement mechanism design in LPDM. Several variants of this mechanism are compared in Tab.~\ref{tab:abl-refine}, including: (1) \textit{No Refine}, where the generated BEV feature is directly passed to the lane decoder without any refinement; (2) \textit{Concat\&FC}, where the concatenated generated and original features are processed with a simple $1 \times 1$ convolution block; (3) \textit{Concat\&ED}, where the concatenated features are processed with an encoder-decoder structure as described in Sec.~\ref{sec:lpdm}; (4) \textit{Add\&FC}, where the generated and original features are added and processed with a $1 \times 1$ convolution block; and (5) \textit{Add\&ED}, where the added features are processed with an encoder-decoder structure. The results show that refining the generated feature improves performance, with the encoder-decoder structure yielding better results than the simpler $1 \times 1$ convolution block. Additionally, feature concatenation and addition lead to similar performance. While cross-attention-based feature fusion could provide a stronger alternative to feature concatenation and addition, we do not explore it here due to GPU memory limitations.
\section{Conclusion}
In this paper, we introduce LaneDiffusion, a novel generative paradigm for centerline graph learning. LaneDiffusion leverages diffusion models to generate lane centerline priors at the Bird's Eye View (BEV) feature level, rather than directly predicting vectorized centerlines. Our approach integrates a Lane Prior Injection Module (LPIM) and a Lane Prior Diffusion Module (LPDM) to effectively construct diffusion targets and manage the diffusion process. Subsequently, vectorized centerlines and topological structures are decoded from these prior-injected BEV features. This modular design renders LaneDiffusion a versatile and efficient add-on for BEV feature–based architectures. Extensive experiments on the nuScenes and Argoverse2 datasets demonstrate that LaneDiffusion achieves state-of-the-art performance in centerline graph learning. Future work will focus on improving the model’s real-time performance and reducing its scale to develop a more lightweight solution.

\section*{Acknowledgment}
This work is supported in part by the National Key R\&D Program of China (2024YFB3908503), in part by the National Natural Science Foundation of China (62322608), and in part by the project of Peng Cheng Laboratory (PCL2025AS214).

\newpage

{
    \small
    \bibliographystyle{ieeenat_fullname}
    \bibliography{main}
}

\end{document}